\documentclass[conference]{IEEEtran}
\IEEEoverridecommandlockouts
\usepackage{cite}
\usepackage{amsmath,amssymb,amsfonts}
\usepackage{algorithmic}
\usepackage{graphicx}
\usepackage{textcomp}
\usepackage{xcolor}

\def\ie{\mbox{\textit{i.e.}}} 
\def\eg{\mbox{\textit{e.g.}}} 

\usepackage[ruled,lined]{algorithm2e}
\usepackage{amsmath}
\usepackage{url}

\def\djy{\textcolor{black}}

\def\lyx{\textcolor{black}}

\def\gjx{\textcolor{black}} 
\newcommand{\gjxx}[2]{\gjx{#2}} 

\def\org{\textcolor{black}} %

\newcommand{\hornet}{the Asian giant hornet } %

\begin{document}

\title{Priority prediction of Asian Hornet sighting report using machine learning methods\\
}

\author{\IEEEauthorblockN{Yixin Liu}
\IEEEauthorblockA{\textit{School of Software Engineering} \\
\textit{South China University of Technology}\\
Guangzhou 510006, PR China}
\and
\IEEEauthorblockN{Jiaxin Guo}
\IEEEauthorblockA{\textit{School of Computer Science and Engineering} \\
\textit{South China University of Technology}\\
Guangzhou 510006, PR China}
\and
\IEEEauthorblockN{Jieyang Dong}
\IEEEauthorblockA{\textit{School of Mathematics} \\
\textit{South China University of Technology}\\
Guangzhou 510006, PR China}
\and
\IEEEauthorblockN{Luoqian Jiang}
\IEEEauthorblockA{\textit{School of Software Engineering} \\
\textit{South China University of Technology}\\
Guangzhou 510006, PR China}
\and
\IEEEauthorblockN{Haoyuan Ouyang}
\IEEEauthorblockA{\textit{School of Software Engineering} \\
\textit{South China University of Technology}\\
Guangzhou 510006, PR China}

\IEEEcompsocitemizethanks{
	\IEEEcompsocthanksitem{Yixin Liu, Luoqian Jiang and Haoyuan Ouyang are with the School of Software Engineering, South China University of Technology. Jiaxin Guo is with the School of Computer Science and Engineering, South China University of Technology. Jieyang Dong is with the School of Mathematics, South China University of Technology.
	E-mail: \{seyixinliu, cs\_guojiaxin\}@mail.scut.edu.cn, 885032560@qq.com, lqrosie728@163.com, 1228194587@qq.com.}
	}
    
}

\maketitle

\begin{abstract}
\lyx{
As infamous invaders to the North American ecosystem, \hornet (\textit{Vespa mandarinia}) is devastating not only to native bee colonies, but also to local apiculture. One of the most effective way to combat the harmful species is to locate and destroy their nests.} \gjx{By mobilizing the public to actively report possible sightings of \hornet, the government could timely send inspectors to confirm and possibly destroy the nests.}
\gjx{However, such confirmation requires lab expertise, where manually checking the reports one by one is extremely consuming of human resources. Further given the limited knowledge of the public about \hornet and the randomness of report submission, only few of the numerous reports proved positive, \ie~existing nests. How to classify or prioritize the reports efficiently and automatically,}
\lyx{
so as to determine the dispatch of personnel, is of great significance to the control of the Asian giant hornet. 
}
\lyx{
In this paper, we propose a method to predict the priority of sighting reports based on machine learning. 
}
\lyx{
We model the problem of optimal prioritization of sighting reports as a problem of classification and prediction. We extracted a variety of rich features in the report: location, time, image(s), and textual description. Based on these characteristics, we propose a classification model based on logistic regression to predict the credibility of a certain report. Furthermore, our model quantifies the impact between reports to get the priority ranking of the reports. 
}
\lyx{Extensive experiments on the public dataset from the WSDA~(the Washington State Department of Agriculture) have proved the effectiveness of our method.}
\end{abstract}

\begin{IEEEkeywords}
\lyx{Sighting Report, Priority Prediction, Machine Learning}
\end{IEEEkeywords}

\section{Introduction}
\djy{Bees are the medium for pollination of plants, and they have made great contributions to human society and ecosystem. Effective pollination by bees can increase the quantity and quality of crops, plant diversity and plant resistance to pests~\cite{steffan2019omnivory}. These crops account for one third of the world's food production, which is an immense contribution. In addition, the products produced by bees can also be further processed into daily necessities for cleaning, beauty and other purposes. Therefore, as contributors to the ecosystem and human society, bees are worthy of protection.} 

However, the bee ecosystem has been severely threatened in recent years\cite{rome2011monitoring}. The Asian giant hornet is a kind of bumblebee with bright yellow feet. As an alien species, they could kill an average of 40 other kinds of bees in their lifetime, which has caused great damage to the local bee ecological environment. At the same time, the hornet is also very aggressive when it comes into contact with humans, \gjxx{and its venom may cause fatal results}{with venom deadly enough to kill}. Therefore, the prevention and control of \hornet is of great significance to the protection of \gjxx{bees and humans}{local apiculture and public safety}. 

Locating the nests of the hornets and destroying \gjxx{its}{them} is one of the most effective way\gjx{s} to control 
\hornet \cite{shams2020detection}.
\djy{Existing methods for \gjxx{the identification of Asian giant hornet nests}{identifying their nests} are mainly based on \gjxx{wireless}{remote} tracking~\cite{tracking} \gjxx{and}{with} image recognition~\cite{shams2020detection}. Although these methods are able to locate nests automatically, they require the equipment of \gjxx{unmanned aerial vehicles}{drones} and face the problem of great search cost.}
In actual situations, the government often mobilizes \gjxx{the masses}{the public} to submit \gjxx{witness}{sighting} reports actively, and then \gjxx{goes}{sends inspectors} to identify and destroy nests in a targeted manner~\cite{pusceddu2019using}. This method can greatly reduce the cost of searching by leveraging the power of citizens. 

However, given that citizens often lack \gjxx{understanding}{related knowledge} of \hornet and the randomness of \gjxx{some citizens' submission of reports}{citizens' report submission}, \gjxx{a large number of sighting reports submitted are generally full of noise}{the report collection is actually full of ``noise''}, and only a small part of the reports \gjxx{sent to inspect}{inspected} are finally confirmed as \gjxx{valid}{positive}. At the same time, the current traditional methods require manual \gjxx{verification of the confidence}{credibility assessment} of the report\gjx{s}, which is time-consuming and inefficient~\cite{zhiquan2021trouble}. To solve these problems, 
under the limitation of human and financial resources, the automatic assessment and \gjxx{sorting}{prioritization} of the \gjxx{confidence of witness }{}reports is an important way to improve the efficiency of government resource utilization and the prevention and control of \gjxx{honeycombs}{pests}. 

\gjxx{The report \gjxx{confidence}{credibility} evaluation field \gjxx{uses}{has used}}{The field of credibility evaluation has used} multi-factor~\cite{gjertson2002multi}, natural language processing \gjx{(NLP)} and other technologies to evaluate the confidence of the report. There are currently multiple application scenarios\gjxx{: Credibility of Bug Report, Credibility of Vehicle Data, Credibility of WEB PAGE}{, \eg~bug report~\cite{olteanu2013web} and vehicle data~\cite{thomas2017assessing}}
. However, there is little research and application of \gjxx{this technology}{these technologies} in the field of \gjxx{identifying the confidence of public witness reports}{sighting report credibility evaluation}.

In this paper, we combine the related technolog\gjxx{y}{ies} of report \gjxx{confidence prediction}{credibility assessment}, and propose a method based on machine learning \gjx{(ML)} to \gjxx{realize}{implement} the automatic \gjxx{prediction of witness report priority}{report prioritization}. Different from \gjxx{the previous}{existing} methods of identifying \gjxx{Asian }{}hornet nests, we locate \gjxx{the hornet nests}{them} from a new perspective based on the \gjxx{confidence prediction}{credibility prediction} of public sighting reports. We abstracted \gjxx{the report priority prediction}{this} problem into a two-classification problem, and extracted a variety of factors in \gjxx{the}{a} report to construct rich features. Finally, we built a classification model based on logistic regression to \gjxx{realize}{implement} automated prediction.

Our contributions are summarized as follows:
\begin{itemize}
    \item We have proposed an automated prediction method for the confidence level of sighting reports of \gjxx{Asian giant bees}{hornets}, which provides a new idea for the identification and precise control of \gjxx{bee nests}{pests}. 
    \item We abstract the modeling of the prioritization of \gjxx{witness}{sightings} reports as a \gjxx{two-category}{two-classification} problem. By \gjxx{extracting}{utilizing} a variety of data from the sighting report, \gjxx{rich features are constructed: review trustworthiness, location trustworthiness, and image trustworthiness}{abundant features have been extracted about the credibility of textual notes, sighting location, and attached image(s)}. 
    \item We \gjxx{have built}{build} a machine learning framework based on logistic regression algorithms, which can take into account multiple factors of the report and predict the \gjxx{confidence of the witness report}{report credibility}. At the same time, for \gjxx{the problem of sample imbalance}{the extreme imbalance problem in the sample}, we propose a weighted loss function. On the public data set \gjxx{of}{from} the Washington Department of Agriculture, a large number of experiments have proved the effectiveness of our method. 
\end{itemize}

\section{Methodology}
\begin{figure*}[thbp!]
    \centering
    \includegraphics[width=1\linewidth]{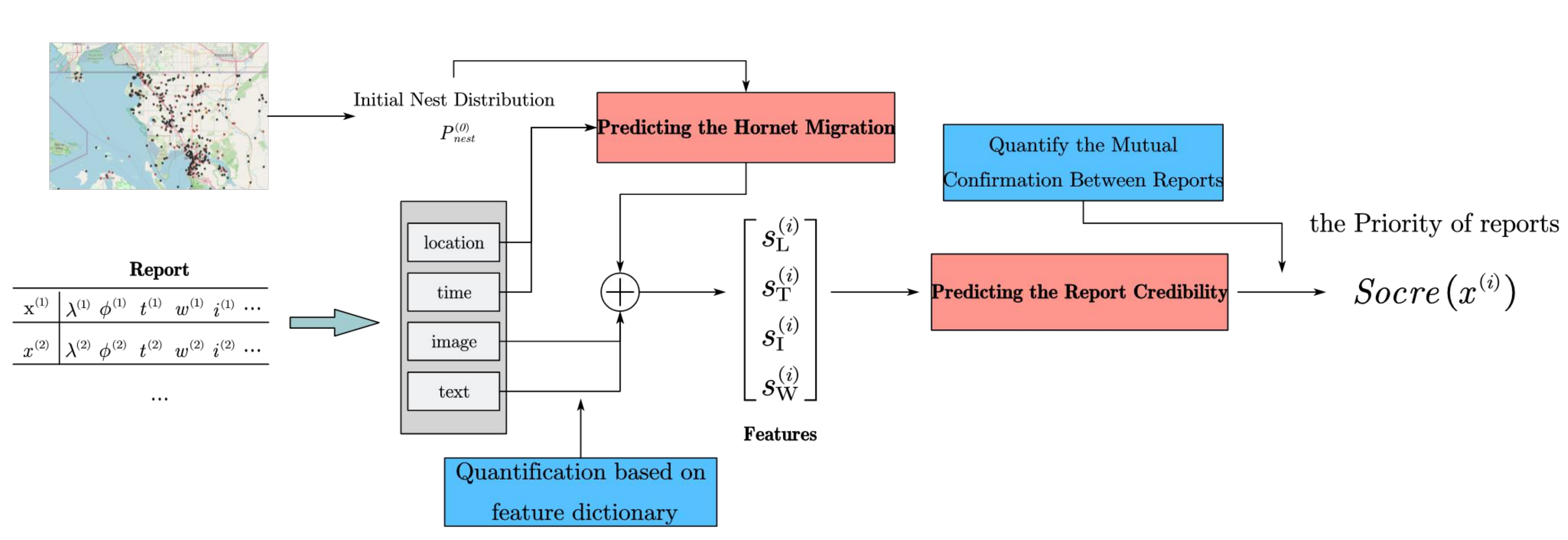}
    \caption{Given a sighting report, we first extract four features: locational, temporal, graphical, and textual. After constructing the feature vector of the report, we use a logistic regression model to predict and classify the credibility of the report. Finally, we quantitatively consider the interaction between the reports and determine the priority of the reports. }
    \label{fig:overview}
\end{figure*}

In this section, we first formalized the problem of priority prediction of sighting reports into a two-category problem. Next, we used multi-factor analysis to construct rich features for the various types of data in the sighting report: location factors, time factors, image factors, and text factors. Finally, we built a machine learning model based on logistic regression to predict the certainty of the report, and further determined the priority of the report based on the confirmation relationship among the reports. We show the overall architecture of the model in Fig.~\ref{fig:overview}. 

\subsection{Problem Definition}
\lyx{To maximize the use of limited government resources, it is essential to accurately predict the credibility of an sighting report so as to prioritize the report, and thus efficiently complete the elimination of the Asian Hornet nest. For a sighting report $x^{\left( i \right)}\sim u\left( \cdot \right)$, where $u\left( \cdot \right) $ denotes a certain distribution that the report data is sampled from. Due to the possibility of mistaken, there will be two different categories. Let the report category label be $y\in{\{0,1\}}$, where $y=1$ denotes that the report confirms the existence of a nest, and $y=0$ means that the report is a false positive. We denotes the probability that report $x^{\left( i \right)}$ belongs to the true positive as $p^{(i)}\in [0,1]$. Given multiple sighting reports, we can be prioritized according to their confidence probability $p^{\left( i \right)}$. For credible reports, we seek to get a higher probability $p^{(i)}$ and vice versa. Formally, we need to solve the following classification optimization problem, that is, to minimize the Cross Entropy in Equation~\ref{target}: }
\begin{equation}
    \max  \mathbb{E}_{x^{\left( i \right)}\sim u\left( \cdot \right)}\left[ y^{\left( i \right)}p^{\left( i \right)}+\left( 1-y^{\left( i \right)} \right) \left( 1-p^{\left( i \right)} \right) \right]
    \label{target}
\end{equation}

\lyx{To solve the above optimization problem, we seek to fit a function $p^{(i)}=f(\cdot|x^{(i)};\theta)$ to predict the confidence probability of the report, so as to further prioritize reports. In the following section, we first construct features from multiple dimensions of the report, and then build a machine learning predictor based on these features to learn an optimal function.}
\subsection{Features Extraction}

\textbf{Location Feature. }
\lyx{For the location factor, we get the probability feature of reporting the hive found at a certain position by constructing the prior distribution probability feature of the Asian Hornet in space based on the mechanism of bee migration. We first briefly explain the habits of the Asian Hornet. The hornet will only be seen by the public from April to December of a year since its will hibernate in winter. During migration season, the average distance between the migration destination and the original nest is about 30km. After arriving at the destination, the bees move around the nest with a maximum radius of 8km. Based on these habits, given the probability distribution of the initial hive in space, we aim to construct the observation probabilities of bees everywhere in the space. }

\lyx{
Given the initial distribution $P_{x,y}^{(0)}$ of the nest in space, we assume that the distance of the new location of nest from the old one follows a Gaussian distribution with an average value of 30 km. We quantify the relation between probability values before and after the migration as follow:
}
\begin{equation}
\label{eqn:nest update}
    P_\text{nest}^{\left( t+1 \right)}\left( \lambda ,\phi \right) =\sum_{\left( \lambda ^{\prime},\phi ^{\prime} \right)}{\frac{1}{\sigma \sqrt{2\pi}}e^{\frac{-\left( d-30 \right) ^2}{2\sigma ^2}}P_\text{nest}^{\left( t \right)}\left( \lambda ^{\prime},\phi ^{\prime} \right)}
\end{equation}
\lyx{where $\left( \lambda ,\phi \right)$ and $\left( \lambda ^{\prime},\phi ^{\prime} \right)$ represent the latitude and longitude of two different coordinates in space, $P_{\mathrm{nest}}^{\left( t \right)}\left( \lambda ,\phi \right) $ denotes the probability of a nest existing at position $(\lambda ,\phi )$ in month $t$, $\sigma$ is a shape parameter for Gaussian distribution, which measures the difference of distances of the newly selected location from the old one, and thus reflects the migration precision. The larger the $\sigma$, the wider the range over which the queen builds her new nest.
}

\lyx{In non-migratory months, we can spot hornets away from near their nest, as its possible location is definite. The further away from the nest, the lower the probability of spotting hornets, and \textit{vice versa}. Formally, according to the nest distribution as defined by Equation \eqref{eqn:nest update}, we can obtain the probability distribution of spotting hornets in the whole space as:}
\begin{equation}
    P_\text{observe}^{\left( t \right)}\left( \lambda ^{\prime},\phi ^{\prime} \right) =\sum_{\left( \lambda ,\phi \right)}{e^{-\beta _1d}P_\text{nest}^{\left( t \right)}\left( \lambda ,\phi \right)}
    \label{eqn:P observe t}
\end{equation}
\lyx{where $P_{\mathrm{observe}}^{\left( t \right)}\left( \lambda ^{\prime},\phi ^{\prime} \right)$ denotes the probability of finding hornets at position $\left( \lambda ^{\prime},\phi ^{\prime} \right)$ in month $t$ and $\beta_1$ is an impact factor that determines the strength of the influence of distance on the probability of spotting a hornet.}

\textbf{Time Feature. }
\lyx{For the time factor, we considered the migration habit of the hornet. \org{Since they are more prevalent amid April-December, the public are more likely to spot them during this time, and the report is more likely to be positive. In December-March, the only surviving individual queens hibernate underground, so it is unlikely to spot a hornet above the ground. The period from April-December each year is hence defined as the {active period}, denoted by $\mathbb{T}$. The temporal feature of each report was characterized as $s_\text{T}$ by the following rules:}}
\begin{equation}
    s_{\mathrm{T}} = \begin{cases}
        c_T, & T \in \mathbb{T} \\
        0, & \text{otherwise}
    \end{cases}
\end{equation}
\lyx{where $c_T$ is the number of positive reports in month $T$ over the historical data. }

\textbf{Image Feature. }
\lyx{For the image factor, we considered the number of images attached to a sighting report for the reason that report with graphical proof improves the credibility of the report. In the process of manual verification, the existence of image information is also an important factor to consider for deciding the sending of personnel. We therefore define $s_\text{I}$ to represent the integrity of image information:}
\begin{equation}
	    s_{\mathrm{I}} = \begin{cases}
        n_\text{I}, & \text{image(s) provided} \\
        0, & \text{otherwise}
    \end{cases}
\end{equation}
\lyx{where $n_\text{I}$ denotes the number of image(s) attached to the report.	}

\textbf{Text Feature. }
\lyx{For the textual factor, we considered the contribution of the length of the text and the keywords contained in the text to the credibility of the report. \org{As for the length of the text, we believe that the length of the text is proportional to its credibility because it contains more evidence. In addition, we extracted two types of text keywords. If the note involves more features closer to \textit{V. mandarinia} than other species, the report is less likely to be mistaken. We constructed a hornet feature dictionary (Table~\ref{tab:The dictionary of key characteristics of Hornet}) from a glossary extracted from attached data, related \textit{Wikipedia} webpages, and literatures. For each report, we are then able to calculate the word frequencies of all the words it contains with respect to this dictionary and obtain its word frequency feature.}}
\gjx{
\begin{table}[tbhp!]
\centering
\caption{The dictionary of key characteristics of \textit{V. mandarinia}}
\label{tab:The dictionary of key characteristics of Hornet}
\resizebox{\linewidth}{!}
{
\begin{tabular}{c|c|c}
\begin{tabular}[c]{@{}c@{}}
\end{tabular} & Asian giant hornet                          & Other confusing hornets                \\ \hline
Nest Location                                                 & \begin{tabular}[c]{@{}c@{}}"Underground","forests",\\ "burrows","roots","trunks", $\cdots$\end{tabular}          & \begin{tabular}[c]{@{}c@{}}"Time limbs", "house eaves",\\  "exposed", "lawns",  $\cdots$\end{tabular} \\ \hline
Body Appearance                                                         & \begin{tabular}[c]{@{}c@{}}"Yellow heads", "black thorax", \\ "striped abdomens", "giant", $\cdots$\end{tabular} & \begin{tabular}[c]{@{}c@{}}"Small",\\ "black and white colored", $\cdots$\end{tabular}     
\end{tabular}
}
\end{table}
}
\lyx{Combining these two points, we define $s_\text{W}$ as the text feature to evaluate a report:}
\begin{equation}
    s_{\mathrm{W}}=\frac{1}{n_{\mathrm{W}}}\sum_{i=1}^{n_{\mathrm{W}}}{\left( q_i-k_{\mathrm{i}} \right)}+\beta _2\log \left( n_{\mathrm{W}}+1 \right) 
\end{equation}
\subsection{Priority Prediction} 
\textbf{Classification model. }
\lyx{Logistic regression is a generalized linear model, which conducts mapping from any real number to a probability values. For a report $x^{\left( i \right)}$ which is characterized by the feature vector $\mathbf{s}$, the logistic function can be used to construct its probability $p$.}
\begin{equation}
    p\left( x^{\left( i \right)} \right) =\frac{1}{1+e^{-\theta ^T\phi \left( x^{(i)} \right)}}
    \label{logistic}
\end{equation}
\lyx{
where $\phi \left( \cdot \right)$ is the feature mapping for reports (yielding $\mathbf{s}^{(i)}$), and $\theta$ is the parameter of the model to learn.
}
\lyx{
To learn the model, we seek to minimize a loss function called \emph{binary cross-entropy function} denoted as $H(\cdot)$. However, there are only few positive report among a large number of reports, which means that dataset is usually extremely unbalanced. To tackle this problem, we use the weighted binary cross-entropy function, where, for the correctly classified fraction, \textit{viz.} when $y=0$, if our model predicts it as misclassification, we scale up its loss by a degree of $\tau$ to ensure that our model would not predict all samples as misclassifications:
}
\begin{equation}
    H\left( p,y \right) =\tau y\log \left( p \right) +\left( 1-y \right) \log \left( 1-p \right)
    \label{eqn:cross entropy}
\end{equation}
\lyx{where $\tau$ is the weighting coefficient of the sample status, set as the reciprocal of the percentage of non-misclassified reports. On this basis, we can obtain the overall loss function $J(\cdot)$ of the given dataset $\mathbf{x}$:}
\begin{equation}
    \min  J\left( \theta \right) =\frac{1}{n}\sum_{i=1}^n{H\left( h_{\mathrm{\theta}}\left( x^{\left( i \right)} \right) ,y^{\left( i \right)} \right)}+\frac{\beta _3}{2}\left\| \theta \right\| _{2}^{2}
\end{equation}
\lyx{where $h_{\theta}(\cdot )$ is the logistic regression function with parameter $\theta$, and $\beta_3$ is the balancing coefficient for regular terms.}

\lyx{We update the parameter of model using stochastic gradient descent method. Let $\eta$ be the step of parameter update, we train our model to minimize the loss $J(\theta)$.}
\begin{equation}
\theta ^{\prime}\gets \theta -\eta \frac{\partial J\left( \theta \right)}{\partial \theta}.
\end{equation}

\textbf{Priority determination. }
\lyx{In the previous section, we predict the credibility of report based on four dimensions of the report content. However, to predict the priority of report, we have to consider their connection. Intuitively, if there are many reports submitted in a certain area at the same time, it highly indicates that this area is likely to have nest hidden. Therefore, based on the forecast confidence of a single report, we seek to further consider the mutual confirmation relationship between the reports, so as to achieve more accurate report priority ranking. }

\lyx{Specifically, the predicted possibility $p$ in Equation~\ref{logistic} only considers one report and does not consider its inter-corroboration with reports that are similar to it in space and time. If the reports are in the same migration cycle~(which means that the reports are between April of a certain year and March of the next year), within which the hornets do not migrate, we are interested in whether the reports lead to the same nest; if different, we consider whether the nests that the reports lead to are related. For this, we defined the mutual influence factor $F_{i,j}$ between reports $x_i,x_j$ as:}
\begin{equation}
\label{eqn:F i j}
    F_{i,j} = \begin{cases}
        e^{-\lambda d_{ij}}, & \text{if }t_i=t_j \\
        \frac{1}{\sigma \sqrt{2\pi}}e^{\frac{-1}{2\sigma ^2}\left( d_{ij}-30\Delta t_{i,j} \right) ^2}, & \text{otherwise}
    \end{cases}
\end{equation}

\lyx{
where $\Delta t_{i,j}$ denotes the difference between the migration cycles of $i$-th and $j$-th report.
}
\lyx{Based on $F_{i,j}$, we construct the priority evaluation $Z$ of a report:}
\begin{equation}
    Z_i=\sum_{j=1}^n{F_{i,j}}p_j
\end{equation}
\lyx{where $Z_i$ denotes the priority of $i$-th report and $n$ denotes the total number of reports involved in ranking. According to Equations \eqref{eqn:F i j} and \eqref{logistic}, the priority $Z_i$ of each report can be obtained. The greater the $Z$-value of a report, the more likely it is to be positive, and thus the sooner it should be investigated.}
\section{Experiments}
\lyx{In the experiment, we used the Asian Hornet sighting report dataset\footnote{The dataset is available at \url{https://agr.wa.gov/departments/insects-pests-and-weeds/insects/hornets/data } } published by the WSDA~(Washington State Department of Agriculture) to verify our model. This dataset includes the sighting report ranging from 2019 to 2021, which details of sighting report and the label after reviewed by the official. }

\lyx{We use accuracy metric to evaluate the classification model and conduct an ablation experiment to demonstrate the effectiveness of our proposed module. }

\subsection{Implementation Details}
\textbf{Data Cleaning. }
\lyx{Among the 4,400 samples, the main informational features of each report were the coordinates, date, text, (possible) images, and officially given identification label. We excluded the following parts of the data: 15 reports with invalid date, 15 with ``unprocessed'' label, and 56 prior to 2019. These parts account for only a small fraction, and contribute little to our later study. After cleaning, we got 4,355 pieces of data. There are three categories of result labels in the cleansed data: positive, negative, and unverified. The number of sample categories is highly unbalanced, with only 14 sightings verified as positive cases, accounting for only 0.3\%, and the remaining half of each case being negative and unverified, respectively.}

\textbf{Parameter Setting. }
\lyx{To determine the initial distribution $P_{x,y}^{(0)}$ in Equation~\ref{eqn:nest update}, we use the confirmed sighting report locations of the data set from March 2019 to April 2020 as the initial nest probability distribution of Washington State. The probability value of each positive reported location is set to 1, and the other locations are all set to 0. The parameter $\beta_1$ in Equation~\ref{eqn:P observe t} is set to $0.57$, and the $\sigma$ in Equation~\ref{eqn:nest update} is set to $10$.}
\subsection{Results of classification prediction}
\lyx{Note that the category of report in this dataset are extremely unbalanced. We conduct an ablation experiments to prove the effect of the weighted loss coefficient in Equation~\ref{eqn:cross entropy}. To this end, we take the parameter $\tau$ as a model parameter, and obtain the training and test sets by 5-fold split~(we will ensure the inclusion of samples of both statuses in each set). Then, we performed grid search for $\tau \in \left\{ \tau _0-\Delta \tau ,\tau _0,\tau _0+\Delta \tau \right\}$, with a 5-fold cross-validation for each parameter (still ensuring both inclusions). Since the reports were mostly negative, the na\"ive use of average accuracy would not evaluate well the prediction of our model, so we took the average accuracy on positive reports as the metric.}
\begin{figure}[htbp!]
    \centering
    \includegraphics[width=1\linewidth]{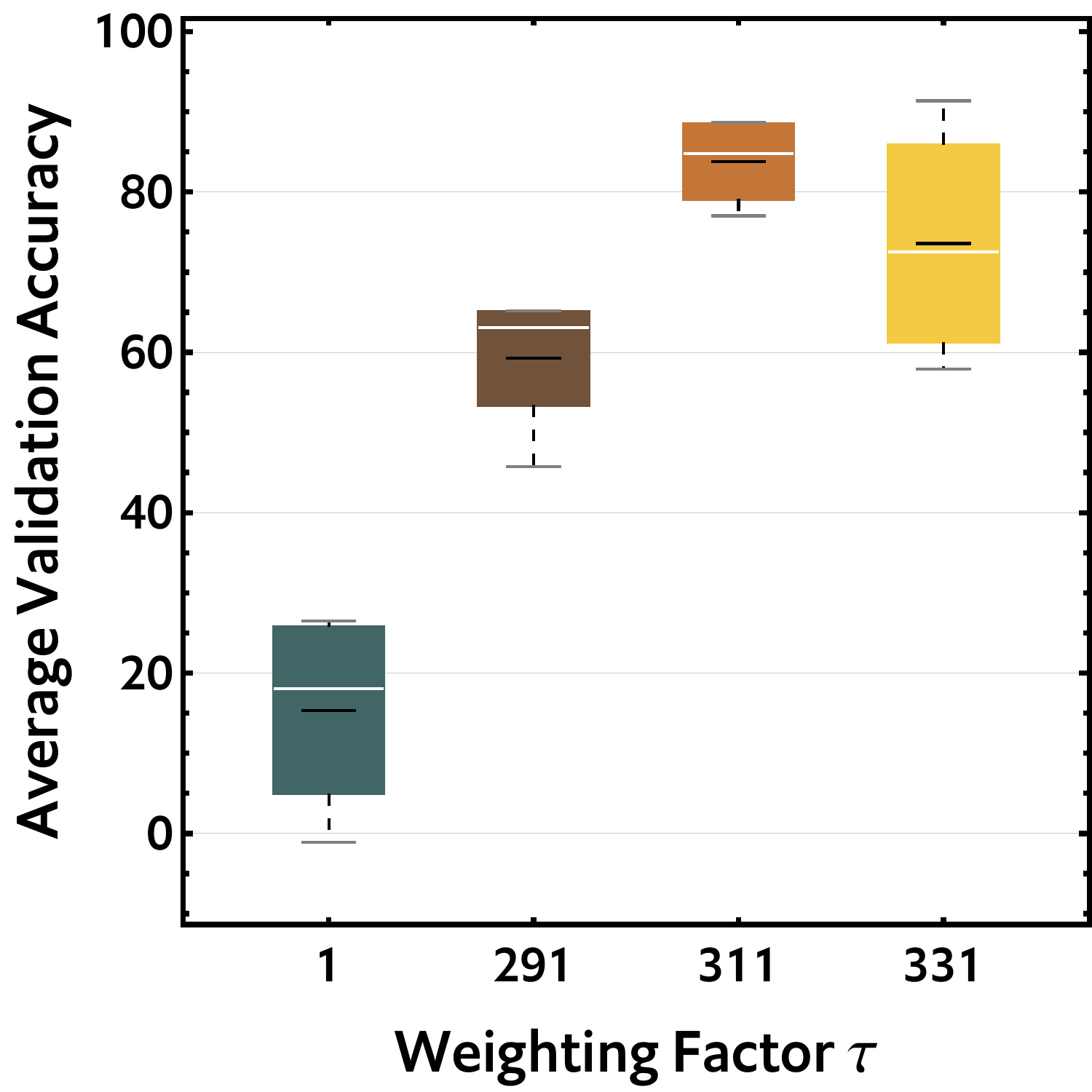}
    \caption{\footnotesize Under different scaling parameter $\tau$ settings, the cross-validation accuracy on report samples of imbalanced statuses.}
    \label{fig:unblance}
\end{figure}
After searching for an optimal parameter setting, a prediction accuracy of $75\%$ is achieved on only few samples. Considering the prediction difficulty due to the extreme imbalance of the samples (with $\text{positive}/\text{negative} \approx 1/400$), the model performs well enough. Also, comparing results with and without setting the weighting parameter ($\tau=1$ for the original binary cross-entropy function), we find that the weighting correction parameter contributes much to improved performance. With the best parameter settings~($\tau=311$), our classification model can achieve the average accuracy of 83.5\%. However, compared with recent related works~(\eg~ \cite{shams2020detection} achieved detection rate of 93\%), there is still room for improvement in our method. 
\subsection{Results of priority prediction}
\begin{figure}[htbp!]
    \centering
    \includegraphics[width=1\linewidth]{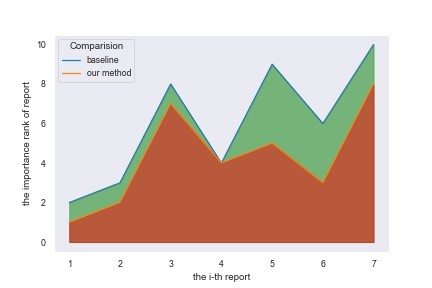}
    \caption{\footnotesize The priority rankings of the $7$ positive samples in the validation set with different ranking models. For these positive reports, the higher the priority predicted by the model, the better. Noted that we use the sort position of the report among $14$ report in testing set to indicate its priority.}
    \label{fig:ranking}
\end{figure}

\lyx{To prove that introducing the mutual influences among reports improves the priority sorting, we performed ablation experiments to demonstrate the effectiveness of our prioritization module by comparing to the baseline method that directly uses the probability computed using Equation~\ref{logistic} to sort the priority. We construct a small and balanced dataset that includes 28 reports, half of which are positive. Specifically, the dataset was split as 1:1, with 14 for training and 14 for testing. }\djy{The results in Fig.~\ref{fig:ranking} show that the priority of positive sample predicted by our method is much higher compared to the baseline, which means that our method can better identify high-priority reports. This demonstrates the effectiveness of the purposed module. }

\section{Conclusion}
\djy{In this paper, we proposed the priority prediction of Asian Hornet sighting reports based on Machine Learning, which solved the problem that prioritizing Asian Hornet sighting reports efficiently and automatically under the limitation of resources. We formalized the problem of priority prediction of sighting reports into a two-category problem. To characterize a sighting report, we construct rich features from four dimensions: location, time, image and text. Then, we purpose a machine learning model based on logistic regression to predict the credibility of the report, and determined their priority based on the relationship among the reports. }
\djy{
Extensive results on the public dataset demonstrate the effectiveness of the proposed method.
}
\bibliography{ref}
\end{document}